\documentclass[11pt,a4paper,twoside]{article}
\usepackage{times}
\usepackage{graphicx}
\begin{document}
%\tableofcontents

%\def\SHORTTITLE  {Data structuring for the ontological modelling of wind energy systems}%

%\vspace*{3cm}
%\antet
%\markboth{\hfill  Adrian Groza}{\hfill \SHORTTITLE}

%\begin{center}
%{\LARGE  \textbf{Data structuring for the ontological modelling of wind energy systems}} %17pt
%\par\vspace*{0.5cm}
%{\textbf{Adrian Groza}}
%\end{center}

%\vspace*{0.75cm}
%%%%%%%%%%%%%%%%%%%%%%%%%%
%%%%%% PROOF.TEX %%%%%%%%%
%%%%%%%%%%%%%%%%%%%%%%%%%%
\tolerance 10000
\newtheorem{theorem}{Theorem}
\newtheorem{lemma}{Lemma}
\newtheorem{definition}{Definition}
\newtheorem{example}{Example}
\newtheorem{xca}{Exercise}
\newtheorem{remark}{Remark}
\newtheorem{proposition}{Proposition}
\newtheorem{corollary}{Corollary}
%Please use these definitions for Latex entities (theorem, lemma, etc.)
%If you need other definitions add to this list and notify us, by e-mail, about this.
%--------------------------------------

\makeatletter
\renewcommand\section{\@startsection {section}{1}{\z@}%
                                   {-3.5ex \@plus -1ex \@minus -.2ex}%
                                   {2.3ex \@plus.2ex}%
                                   {\normalfont\fontsize{16pt}{14pt}\selectfont\bfseries}}
\renewcommand\subsection{\@startsection{subsection}{2}{\z@}%
                                     {-3.25ex\@plus -1ex \@minus -.2ex}%
                                     {1.5ex \@plus .2ex}%
                                     {\normalfont\fontsize{13pt}{13pt}\selectfont\bfseries}}
\renewcommand\subsubsection{\@startsection{subsubsection}{3}{\z@}%
                                     {-3.25ex\@plus -1ex \@minus -.2ex}%
                                     {1.5ex \@plus .2ex}%
                                     {\normalfont\fontsize{11pt}{11pt}\selectfont\bfseries}}
%\makeatother

\title{Data structuring for the ontological modelling of wind energy systems}
\date{}
\maketitle
\begin{abstract}
Small wind projects encounter difficulties to be efficiently deployed, partly because wrong way data and information are managed.
Ontologies can overcome the drawbacks of partially available, noisy, inconsistent, and heterogeneous data sources, by providing a semantic middleware between low level data and more general knowledge.
In this paper, we engineer an ontology for the wind energy domain using description logic as technical instrumentation. 
We aim to integrate corpus of heterogeneous knowledge, both digital and human, in order to help the interested user to speed-up the initialization of a small-scale wind project. 
We exemplify one use case scenario of our ontology, that consists of automatically checking whether a planned wind project is compliant or not with the active regulations.

\end{abstract}
% Next you must introduce the contents of your article

%=====================================================
% 1. Introduction and Motivations
\section{Introduction}
\label{sec:intro}
%=====================================================
Small wind projects in the sector of renewable energies  
encounter difficulties to be efficiently deployed, partly because the wrong way data and
information are managed~\cite{lamanna2014renewable,octavian2012online}.
Ontologies can overcome the drawbacks of partially available, noisy, inconsistent, and heterogeneous data sources~\cite{nguyen2013proactive}.
Domain specific ontologies have already been developed for the renewable energy sector~\cite{Nguyen2013150,kayed2013renewable,nguyen2014offshore}.

In this line, we aim to develop an ontology for wind energy domain. 
The ontology was developed in the RacerPro (Renamed ABox and Concept Expression Reasoner Professional) knowledge representation and reasoning system~\cite{HHMW12}. 
In our view, RacerPro and the corresponding Knowledge Representation System Specification (KRSS) syntax for Description Logic axioms are powerful technical instrumentation that support ontology engineering behind the basic capabilities provided by GUI-based ontology editors. 
Compared to the ontologies listed in Table~\ref{tab:ontologies}, the particularity of our ontology is to complement knowledge extracted from various maps with local observations about the  location. 
Aiming to deal with small wind projects, we rely on 30-50 meter height wind maps and  
we also include community-scale wind resource maps to quantify the wind resource.

\begin{table}
\begin{center}
\caption{Examples of ontology-based systems in the renewable energy sector.}
\begin{footnotesize}
 \begin{tabular}{|l|l|p{13cm}|}\hline
\textit{Ontology} & \textit{Ref.} & \textit{Short description} \\ \hline
OpenWatt &  \cite{lamanna2014renewable} & Global schema for data about solar energy, wind energy, and biomasses. \\ \hline
EMA  & \cite{Harith2013}& RDF/OWL model of the French electricity company. The ontology-based Energy Management Adviser (EMA) provides personalised tips for 300,000 clients of the company. \\ \hline
WONT  & \cite{Kucuk2014484} & Semi-automatic created ontology from the Wiki articles in the domain of wind energy (http://www.ceng.metu.edu.tr/we120329/wont.owl). \\ \hline
TurbMon  & \cite{papadopoulos2009wind}& Ontology for wind turbines' condition monitoring: focus on wind turbine components and fault detection by means of SPARQL queries. \\ \hline 
SEMANCO & \cite{corrado2015data} & Ontology for urban planning and energy management that describes regions, cities, and buildings; energy consumption and CO2 emission indicators (based on ISO/CD 16346, ISO/CD 16343), climate and socio-economic factors that influence energy consumption.\\ \hline
\end{tabular}
\end{footnotesize}
\end{center}
\label{tab:ontologies}
\end{table}

The remaining of the paper is structured as follows: 
Section~\ref{sec:technical} briefly introduces the KRSS syntax in which the ontology was developed. 
Section~\ref{sec:related} illustrates how the available knowledge was reused. 
Section~\ref{sec:engineer} shows the main engineering steps of the ontology.
Section~\ref{sec:queries} depicts how the ontology can be interrogated, while section~\ref{sec:conclusions} concludes the paper. 

%Specifically, the KRSS syntax facilitates the speed of writing axioms, while RacerPro provides a wide set of primitives to introduce concepts, roles, constraints, debugging axioms and query the knowledge base.    

\section{Technical instrumentation}
\label{sec:technical}
%\subsection{Open Linked Data}

%\subsection{Description Logic}
The wind data is modelled in an ontology, which is a semantic framework for organising information. 
We formalise the wind ontology in Description Logic (DL). 
In the description logic $\mathcal{ALC}$, concepts are built using the set of constructors formed by
negation, conjunction, disjunction, value restriction, and existential restriction~\cite{baader2003description}, as shown in Table~\ref{tab:dl}. 
Here, $C$ and $D$ represent concept descriptions, while $r$ is a role name.
The semantics is defined based on an interpretation $I = (\Delta^{I}, \cdotp^{I})$, where the domain $\Delta^{I}$ of $I$ contains a non-empty set of individuals, and the interpretation function $\mathit{\cdotp^I}$ maps each concept name $C$ to a set of individuals $C^I\in \Delta^I$ and each role $r$ to a binary relation $r^I\in \Delta^I \times \Delta^I$. 
The last column of Table~\ref{tab:dl} shows the extension of $\cdotp^{I}$ for non-atomic concepts. 

\begin{table}
\centering
%\begin{footnotesize}
\caption{KRSS syntax and semantics of $\mathcal{ALC}$ description logic.}
\begin{footnotesize}
\begin{tabular}{|l|l|l|}
\hline
{\it Constructor} & {\it Syntax} & {\it Semantics} \\
\hline
%negation & $\mathit{\neg C}$ & $\mathit{\Delta^{I} \setminus C^I}$\\ \hline 
negation & \texttt{(not C)} & $\Delta^{I} \setminus C^I$\\ \hline 
conjunction & \texttt{(and C D)} & $C^I \cap D^I$\\ \hline
disjunction & \texttt{(or C D)} & $C^I \cup D^I$\\ \hline
existential restriction & \texttt{(some r C)} & $\{x\in \Delta^I | \exists y: (x,y)\in r^I \wedge y\in C^I\}$\\ \hline
value restriction & \texttt{(all r C)} & $\{x\in \Delta^I | \forall y: (x,y)\in r^I \rightarrow y\in C^I\}$\\ \hline 
individual assertion & \texttt{(instance a C)} & $\{a\} \in C^I$\\ \hline
role assertion & \texttt{(related a b r)} & $(a^I,b^I) \in r^I$\\ \hline
\end{tabular}
%\end{footnotesize}
\end{footnotesize}
\label{tab:dl}
\end{table}

An ontology consists of terminologies (or TBoxes) and assertions (or ABoxes).
A terminology $\mathit{TBox}$ is a finite set of terminological axioms of the form \texttt{(equiv C D)} or \texttt{(implies C D)}.
An assertional box $\mathit{ABox}$ is a finite set of concept assertions \texttt{(instance i C)}, role assertions \texttt{(related i j r)}, or attribute fillers \texttt{(attribute-filler i value a)}, where $C$ designates a concept, $r$ a role, $a$ an attribute, and $i$,$j$ are two individuals.
Usually, the unique name assumption holds within the same $\mathit{ABox}$. 
%\begin{definition}
A concept $C$ is satisfied if there exists an interpretation $I$ such that $C^I \neq \emptyset$.  
%The concept $D$ subsumes the concept $C$ ($\mathit{C\sqsubseteq D}$) if $\mathit{C^I \subseteq D^I}$ for all interpretations $I$.
The concept $D$ subsumes the concept $C$, represented by \texttt{(implies C D)} if $\mathit{C^I \subseteq D^I}$ for all interpretations $I$.
%\end{definition}
Constraints on concepts (i.e. \texttt{disjoint}) or on roles 
(\texttt{domain}, \texttt{range} of a role, \texttt{inverse} roles, or \texttt{transitive} properties) can be specified in more expressive description logics. 
We provide only some basic terminologies of DL in this paper to make it self-contained. 
For a detailed explanation about families of description logics, the reader is referred to~\cite{baader2003description}.

\section{Reusing related ontologies}
\label{sec:related}
The process of engineering the ontology was started by:  
1) specifying use cases of the ontology, 
2) defining a set of competency questions, and 
3) analysing the existing ontologies for possible reuse.

First, the {\it use cases} of the ontology include 
i) "the assessment of the feasibility to install a particular small-scale wind turbine in a given location" or 
ii) ``checking whether a small wind project is compliant with current regulations. 
%A usage scenario would be: \textit{``A homeowner wants to install a small wind turbine for his cottage. He wants to connect the wind turbine to the transmission lines of the local electricity company''}.

Second, a solution to narrow the scope of an ontology is to start by defining a list of {\it competency questions} (CQs)~\cite{noy2001ontology}.
CQs are questions that an ontology should be able to answer in order to satisfy use cases. 
Thereby, CQs represent initial requirements and they can be used to validate the ontology. 
Having the role of a requirement, each CQs are initially written in natural language (see Table~\ref{tab:cq}).
Then, CQs are formalised in the new Racer Query Language (nRQL)~\cite{HHMW12} for the task of ontology validation, .

\begin{table}
\begin{center}
\caption{Sample of competency questions for the wind turbine domain.}
\begin{footnotesize}
 \begin{tabular}{|l|l|}\hline
$CQ_1$ & Which is the most adequate wind turbine class for a given location?\\ \hline
$CQ_2$ & Is it norm-compliant to install a wind turbine of a specific class in a given location?\\ \hline
$CQ_3$ & Which is the wind rose distribution for a specific location?\\ \hline
$CQ_4$ & What type of vegetation does exist within a radius of 200 meters? \\ \hline
$CQ_5$ & Which wind turbine types are installed nearby?\\ \hline
$CQ_6$ & Which turbines have alarms for the generator component occurred after a maintenance activity? \\ \hline
\end{tabular}
\end{footnotesize}
\end{center}
\label{tab:cq}
\end{table}

\begin{table}
\caption{Reusing related ontologies.}
\begin{footnotesize}
 \begin{tabular}{|l|l|l|}\hline
 \textit{Ontology} & \textit{URI} & \textit{Short description}\\ \hline
GeoNames & http://www.geonames.org/ & Contains 8.3 million geonames toponym. \\ \hline
BayesOWL & http://semanticweb.org/wiki/Bayes\_OWL & A probabilistic extension to the Ontology Language OWL.\\ \hline
Sensor & http://www.w3.org/2005/Incubator/ssn/ssnx/ssn & Describes various sensors and observations.\\ \hline   
\end{tabular}
\end{footnotesize}
\label{tab:related}
\end{table}
Third, several types of ontologies are needed to merge the input data which is collected in different format and several measurements types are exploited. 
Thus, existing knowledge bases like wind ontology or measurement ontologies can be exploited. 
For handling data directly coming from sensors, knowledge about the type of sensor or error and transmission rate are needed, which can be extracted from a specific sensor ontology. Merging different GIS maps requires geographical knowledge. 
%An analysis is needed to check if these sources of knowledge meet the requirements or they will be re-used in a larger ontology designed for the wind potential domain. 
The reused ontologies are listed in Table~\ref{tab:related}.
First, from the GeoNames ontology we exploited: 
i) hypsographic features (concepts like \texttt{Mountain, Hill, Valley, Slope, Mesa, Cliff}),
ii) road features (concept like \texttt{Junction}, or roles like \texttt{roadWidth}),
iii) populated places (\texttt{City, Area, ResidentialArea}), or 
iv) vegetation features (\texttt{Forrest, Orchard, Scrubland, Vineyard}).
Second, we used BayesOWL ontology~\cite{ding2006bayesowl} to represent random variables such as \texttt{WindRose} or \texttt{WindShear}. 
Third, we reused concepts from the sensor ontology to describe knowledge about wind-related sensors, as in the following definitions:

\begin{footnotesize} 
\begin{verbatim}
(implies (or Anemometer WindProfiler WindVane) Sensor)
(implies (or CupAnemometer PropellerAnemometer SonicAnemometer) Anemometer)
\end{verbatim}
\end{footnotesize}

\section{Engineering the wind potential assessment ontology}
\label{sec:engineer}
To develop the wind ontology, we follow the methodology in~\cite{noy2001ontology} and we also enact various ontology design patterns~\cite{pollockontology,groza2014ontology}. 
The ontology is a modular one, consisting of a core formalisation and T-boxes for modeling various aspects in the wind energy domain: sub-components of a turbine, classes of turbines, potential of a location, etc.

The sub-componets of a wind turbine are represented by the transitive role \texttt{hasPart}:
\begin{footnotesize} 
\begin{verbatim}
(define-primitive-role hasPart :transitive t)
\end{verbatim}
\end{footnotesize}

A turbine has one base, one tower, one nacelle, and several blades:

\begin{footnotesize} 
\begin{verbatim}
(implies WindTurbine (and (=1 hasPart Base) 
                          (=1 hasPart Tower) 
                          (=1 hasPart Nacelle) 
                          (some hasPart Blade)))
\end{verbatim}
\end{footnotesize}

Given the transitivity of the role \texttt{hasPart},
the system is able to deduce all the sub-components of a turbine. 
The nacelle houses a gearbox and generator, which can be either variable or fixed speed:

\begin{footnotesize} 
\begin{verbatim}
(implies Nacelle (and (=1 hasPart Gearbox) (=1 hasPart Generator)))
(implies Generator (or VariableGenerator FixedSpeedGeneartor))
(disjoint VariableGenerator FixedSpeedGeneartor)
\end{verbatim}
\end{footnotesize}

Component prices and technical specification are attached to instances of the concept \texttt{WindTurbine}:

\begin{footnotesize} 
\begin{verbatim}
(instance whisperH20 (and WindTurbine (= 595 hasPrice)))
(instance towerKit1 (and Tower (= 450 hasPrice))) 
(related whisperH20 towerKit1 hasPart)
(attribute-filler whisperH20 20 hasBladeArea) 
\end{verbatim}
\end{footnotesize}

Small wind turbines, defined as having a swept area less than 200 $m^2$, are usually installed between 15 and 40 m high: 

\begin{footnotesize}
\begin{verbatim}
(implies SmallWindTurbine (and WindTurbine 
                               (min sweptArea 200) 
                               (min high 15) 
                               (max high 40)))
\end{verbatim}
\end{footnotesize}

%\subsection{Populating the ontology}
A good wind resource is one where wind speeds average 16 mph or more over the course of a year:

\begin{footnotesize}
\begin{verbatim}
(define-concrete-domain-attrbiute speedAverage :domain Location :type integer)
(implies GoodWindResource (and WindResource (min speedAverage 16)))
\end{verbatim}
\end{footnotesize}

Wind speeds are categorized by class, from a low of class 1 to a high of class 7.  Wind speeds of class 4 or greater are used for wind power production:

\begin{footnotesize}
\begin{verbatim}
(implies GoodWindResource (or WindClass4 WindClass5 WindClass6 WindClass7)))
\end{verbatim}
\end{footnotesize}

The power of the wind is measured in watts per square meter, and this increases by the cube of the wind speed:

\begin{footnotesize}
\begin{verbatim}
(implies top (= windpower (* windspeed windspeed windspeed))) 
\end{verbatim}
\end{footnotesize}

GIS maps are the main source of data fed in the Aboxes of the ontology: 
(1) wind potential map,
(2) wind power map,
(3) topographical map,
(4) open street map (OSM),
(5) transmission lines,
(6) archaeological map,
(7) vegetation map,
(8) rivers and lakes map,
(9) digital elevation models.
For instance, for converting OSM into KRSS syntax we developed a Java-based converter based on the Osmosis API to import facts about roads.
From the wind power map, we defined the potential of a particular wind resource in three classes for the wind speed measured at 50m: marginal, promising, or excellent: 

\begin{footnotesize}
\begin{verbatim}
(implies MarginalPotentialat50 (and (= hasMeasuredHeight 50) (< 7.5 windSpeed))) 
(implies PromisingPotentialat50 (and (= hasMeasuredHeight 50) (> 7.5 windSpeed) 
                                     (< 10 windSpeed))) 
(implies ExcellentPotentialat50 (and (= hasMeasuredHeight 50) (> 10 windSpeed))) 
\end{verbatim}
\end{footnotesize}

The general concept \texttt{MarginalPotential} is the union of the marginal potentials at different heigths:

\begin{footnotesize}
\begin{verbatim}
(equivalent MarginalPotential (or MarginalPotentialat15 
                                  MarginalPotentialat50 
                                  MarginalPotentialat100)) 
\end{verbatim}
\end{footnotesize}

These concepts can be used to make rough production estimates at sites, given that the accuracy is +/- 10 to 15\% what the actual winds at a site may be.

Access roads of at least 4m wide are assumed necessary, given by:

\begin{footnotesize}
\begin{verbatim}
1. (define-concrete-domain-attribute width :domain Road :type real)
2. (implies TurbineAccessRoad (and Road (>= width 4.0)))
\end{verbatim}
\end{footnotesize}

Wind turbines are efficient in coast, hills, and mountains regions: 

\begin{footnotesize}
\begin{verbatim}
3. (implies WindTurbine (all efficientIn (or Coast Hill Mountain)))
\end{verbatim}
\end{footnotesize}

Theoretically, wind turbines can extract up to 59\% from the energy which passes through it.
Practically, an efficient turbine extracts around 40\% from the wind
potential~\cite{octavian2012online}. 
The individual \texttt{wt1} is an instance of the concept \texttt{WindTurbine} that is able to extract 38\% from the wind energy:

\begin{footnotesize}
\begin{verbatim}
4. (define-concrete-domain-attribute extracts :domain WindTurbine :type real)
5. (instance teoreticalLimit (= extracts 0.59))
6. (instance practicalLimit (< extracts 0.40)) 
7. (instance wt1 (and WindTurbine (= practicalLimit 38)))
\end{verbatim}
\end{footnotesize}

%\begin{figure}
%\caption{Knowledge about wind turbine location.}
%\label{fig:location}
%\end{figure}

%In Fig.~\ref{fig:location} 
The transitive role \texttt{isLocated} connects instances of type \texttt{Entity} with instances of the concept \texttt{Location} (line 21). 
"Partition ontology design pattern"~\cite{pollockontology} is used to define various types of locations (lines 22-23). 
The attributes \texttt{hasLatitude} and \texttt{hasLongitute} are introduced to define a point in space (lines 25-26). 
Because GIS maps are usually achieved for small regions, we defined aboxes for each region (line 27). 
Axiom 29 connects the wind turbine \texttt{wt1} with the point \texttt{p1}. 
Because the role \texttt{isLocated} is transitive the system is able to infer that \texttt{wp1} is located in all concepts representing more general locations than \texttt{p1} (i.e. Dobrogea, Romania, given that \texttt{(related Dobrogea Romania isLocated)}. 

\begin{footnotesize}
\begin{verbatim}
21. (define-primitive-role isLocated :domain Entity :range Location :transitive t)
22. (implies (or Point Area) Location)
23. (disjoint Area Location)
24. (implies IndustrialArea (and Area (some has IndustrialActivity)))  
25. (define-concrete-domain-attribute hasLatitude :type real)
26. (define-concrete-domain-attribute hasLongitude :type real)
27. (init-abox dobrogea-wind-assesment)
28. (instance p1 (and Point (= hasLatitude 44.56) (= hasLongitude 27.54)))
29. (related wt1 p1 isLocated)
\end{verbatim}
\end{footnotesize}

When setting a home-sized wind turbine, the rule of thumb states that turbine's rotor should be at least 10 meters above anything within 150 meters of the tower:

\begin{footnotesize}
\begin{verbatim}
(define-rule (?wt ProperWindTurbine) (and (?wt WindTurbine) 
                                          (?wt ?h1 hasHeight)
                                          (?x top) (?x ?h2 hasHeight)
                                          (?dist DistanceBetween2Objects)
                                          (?dist ?wt hasObject)
                                          (?dist ?x hasObject)
                                          (?dist ?distance hasDistance) 
                                          (< distance 150)
                                          (> ?h1 (+ h2 10)))                                  
\end{verbatim}
\end{footnotesize}

Here, the wind turbine \texttt{?wt} is proper if its heigth \texttt{?h1} is greater then the height \texttt{h2} of any object \texttt{?x} in the ontology \texttt{(?x top)}. 
We enacted the "n-ary ontology design pattern"~\cite{pollockontology} to store the distance between two objects: 

\begin{footnotesize}
\begin{verbatim}
(equivalent DistanceBetween2Objects (and (=2 hasObject) (=1 hasDistance)))
(define-concrete-domain-attribute hasDistance :domain DistanceBetween2Objects :type real)
\end{verbatim}
\end{footnotesize}

The role \texttt{hasWindShear} is used to describe the differences in wind speed at two different heights (line 12) %in Fig.~\ref{fig:parameters}). 
In general, turbulence decreases and wind speed increases as height increases.            
A \texttt{WindRose} (line 13) shows the direction that the wind blows and the frequency of that direction at a particular location. 

%\begin{figure}

\begin{footnotesize}
\begin{verbatim}
11. (define-concrete-domain-attribute hasAverageWindSpeed :domain Location :type real)
12. (define-concrete-domain-attribute hasWindShear :domain Location :type real)
13. (implies WindRose (and (=1 hasDirection Direction) (=1 hasFrequency Frequency)))
14. (instance wr (and (= NW hasDirection) (= hasFrequency 0.6)))
\end{verbatim}
\end{footnotesize}
%\caption{Wind-related knowledge about a specific point.}
%\label{fig:parameters}
%\end{figure}

To represent the random variable \texttt{WindRose} given the variable \texttt{Day} ($P(WindRose|Day)$) we enact the \texttt{ConditionalProbability} concept from the BayesOWL ontology~\cite{ding2006bayesowl}:% (see Fig.~\ref{fig:bayes}). 

%\begin{figure}
\begin{footnotesize}
\begin{verbatim}
41. (define-primitive-role hasCondition :domain ConditionalProbability)
42. (define-primitive-role hasVariable :domain ConditionalProbability)
43. (define-concrete-domain-attribute hasProbabilityValue :domain ConditionalProbability 
                                                          :type real)
44. (instance cp1 bayesOWL:ConditionalProbability)        
45. (instance wr1 (and WindRose RandomVariable))           
46. (instance 27June Day)                                 
47. (related cp1 27June bayesOWL:hasCondition)  
48. (related cp1 wr1 bayesOWL:hasVariable)    
49. (attribute-value cp1 0.6 bayesOWL:hasProbabilityValue)  
\end{verbatim}
\end{footnotesize}

\section{Retrieving information from Aboxes}
\label{sec:queries}

Once a user's situation has been fully described using DL assertions, the RacerPro~\cite{HHMW12} reasoner was used to classify the location as an instance
of one or more of the situation concepts described in the ontology. 

Ontology reasoning can be used to check if the location of a wind turbine does not breach active regulations. 
The following axioms checks whether the wind turbine is located at a minimum distance of 300m from a residential area.  

\begin{footnotesize}
\begin{verbatim}
31. (instance v1 Village)                
32. (implies Village (and geonames:PopulatedPlace geonames:ResidentialArea))                 
33. (instance a1 (and Area (= hasContourPoint p2) (= hasCountourPoint p3) 
                           (= hasContourPoint p4))) 
34. (related v1 a1 isLocated)
35. (define-concrete-domain-attribute hasValue :domain Distance :type integer)
36. (instance d1 (and Distance (= hasObject wt1) (= hasObject v1)))
37. (attribute-filler d1 280 hasValue)
38. (instance minimumDistance (= distance 300))
\end{verbatim}
\end{footnotesize}
%\caption{Knowledge used for geospatial reasoning to check that the location of a wind turbine is norm-compliant.}
%\label{fig:distance}
%\end{figure}
Given the village \texttt{v1} (line 31), we use axioms from GeoNames ontology (axiom 32) to deduce that a village is a residential area. 
We enact AllegroGraph~\cite{aasman2006allegro} to perform geospatial reasoning on various tasks, such as computing the distance between a point and a geographic area. 
Using also the Harversine formula, the minimum distance \texttt{d1} between potential wind turbine location \texttt{p1} and residential area \texttt{a1} of \texttt{v1} (line 34) is asserted in the ontology (line 37). 
Here, area \texttt{a1} is a triangular shape defined by three points (line 33) and \texttt{d1} is an instance of the concept \texttt{Distance} between the starting point the turbine \texttt{wp1} and the ending point the village \texttt{v1} (line 36).
%Note that we needed to employ the "n-ary relation ontology design pattern"~\cite{pollockontology} to store in the distances between several entities.                          
Given that the minimum distance between a turbine an a residential area should be 300m (fact 38) the ontology is able to infer that the turbine \texttt{v1} breaches the above normative condition.

%Fig.~\ref{fig:consistency} illustrates 
Four types of queries on the wind-turbines ontology are illustrated:
i) checking the  ontology consistency (line 71), 
ii) retrieving information about individual \texttt{wt1} (line 72), 
iii) identifing the sub-concepts of the various concepts in the ontology (lines 73-74) and 
iv) retrieving all wind turbines located in Dobrogea region (line 75).

%\begin{figure}
\begin{footnotesize}
\begin{verbatim}
71. (tbox-cyclic?)  (tbox-coherent?) (abox-consistent?)
72. (describe-individual wt1)
73. (concept-children WindTurbine) 
74. (concept-descendents Wind Turbine)
75. (concept-instances (and WindTurbine (some isLocated Dobrogea)))
\end{verbatim}
\end{footnotesize}
%\caption{Checking consistency and retrieving information.}
%\label{fig:consistency}
%\end{figure}

%=====================================================
\section{Conclusion and ongoing work}
\label{sec:conclusions}
%=====================================================
The proposed ontology captures knowledge in the wind energy domain. 
The knowledge was formalised in Description Logic that provides a rich, flexible and fully declarative language for modelling knowledge about environment. 
The specific reasoning services on DL were exploited to deduce new knowledge and to classify a location according to its wind potential.
The ontology can assist the wind farm operators lowering costs and complying with current regulations in the wind energy sector.
The wind ontology will help wind farm operators make informed, evidence based decisions about deployment and maintanance of a wind project. 
%We developed an ontology for supporting reasoning about the wind potential for a specified location. 
%Site-specific measurements using anemometers are considered the most reliable estimates. However, anemometers are costly and require from one to several years to complete. In our approach we included observations of the user in order to increase the estimation provided by small resolution wind power maps. 

%\cite{wimmler2015multi}
We did not focus here on the problem of reasoning on data streams~\cite{groza2012plausible} continously collected from sensors. Ongoing work regards extending the ontology with terrain analysis, infrastructure, environment and complete legal constraints, into a single predictive map that shows the most suitable site to explore for wind energy.

%If there is only one author of the paper, please put blank space for the information of the second author.
%The information of the second author is situated, on each line, before the sign &

% If there are 3 authors of the paper, please replace complete the part given before, for 2 authors,
% with the following part given for 3 authors, without the sign %, from the beginning of each line.

%\vspace*{1cm} {\footnotesize
%\begin{tabular*}{16cm}{p{5.6cm}p{5.6cm}p{5.6cm}}
%Tom& Jerry & Third author complete name\\
% Name of institution 1 & Name of institution 2& Name of institution 3\\
%Name of faculty or department1& Name of faculty or department2 & Name of faculty or department3\\
%Address of institution 1 & Address of institution 2 & Address of institution 3\\
%COUNTRY 1& COUNTRY 2& COUNTRY 3\\
% E-mail: \ {\it name1@server1}& E-mail: \ {\it  name2@server2}& E-mail: \ {\it  name3@server3}
%\end{tabular*}}

\bibliographystyle{plain} 
\bibliography{bib,papers}

\vspace*{1cm} {\footnotesize
\begin{tabular*}{16cm}{p{8.2cm}p{8.2cm}}
Adrian Groza&   \\
Intelligent Systems Group &\\
Department of Computer Science &  \\
Technical University of Cluj-Napoca & \\
Baritiu 26-28, Cluj-Napoca, ROMANIA & \\
E-mail: \ {\it adrian.groza@cs.utcluj.ro}&
\end{tabular*}}

\end{document}